\documentclass[conference]{IEEEtran}
\IEEEoverridecommandlockouts

\usepackage{cite}
\usepackage{amsmath}
\usepackage{mathtools}
\usepackage{graphicx}
\usepackage[english]{babel}
\usepackage[unicode=true]{hyperref}
\usepackage{amssymb}
\usepackage{algorithmic}
\usepackage{graphicx}
\usepackage{textcomp}
\usepackage{tabularx}  
\usepackage{multirow}
\usepackage{url}
\usepackage{wasysym}
\usepackage{array}
\usepackage{marvosym}
\usepackage{hyperref}
\usepackage{flushend}
\usepackage{enumitem}
\usepackage{booktabs}

\usepackage{adjustbox}

\begin{document}
\title{
Enhancing Complex Formula Recognition with Hierarchical Detail-Focused Network
\thanks{*Corresponding author. Email: yujunhui3@360.cn}

}

\author{\IEEEauthorblockN{
Jiale Wang\IEEEauthorrefmark{2}\IEEEauthorrefmark{3},
Junhui Yu\IEEEauthorrefmark{2}\IEEEauthorrefmark{1},
Huanyong Liu\IEEEauthorrefmark{2},
Chenanran Kong\IEEEauthorrefmark{4}
}
\IEEEauthorblockA{
\small
\IEEEauthorrefmark{2} 360 AI Research Institute, China 
\IEEEauthorrefmark{3} Nanyang Technological University, Singapore
\IEEEauthorrefmark{4} Chinese University of Hong Kong (Shenzhen), China
}
}

\maketitle

\begin{abstract}
Hierarchical and complex Mathematical Expression Recognition (MER) is challenging due to multiple possible interpretations of a formula, complicating both parsing and evaluation. In this paper, we introduce the Hierarchical Detail-Focused Recognition dataset (HDR), the first dataset specifically designed to address these issues. It consists of a large-scale training set, HDR-100M, offering an unprecedented scale and diversity with one hundred million training instances. And the test set, HDR-Test, includes multiple interpretations of complex hierarchical formulas for comprehensive model performance evaluation. Additionally, the parsing of complex formulas often suffers from errors in fine-grained details. To address this, we propose the Hierarchical Detail-Focused Recognition Network (HDNet), an innovative framework that incorporates a hierarchical sub-formula module, focusing on the precise handling of formula details, thereby significantly enhancing MER performance. Experimental results demonstrate that HDNet outperforms existing MER models across various datasets.
\end{abstract} 


\section{Introduction}
\label{sec:intro}

\begin{figure}[htb]
    \centering
    \includegraphics[width=0.8\linewidth, height=2.2cm]{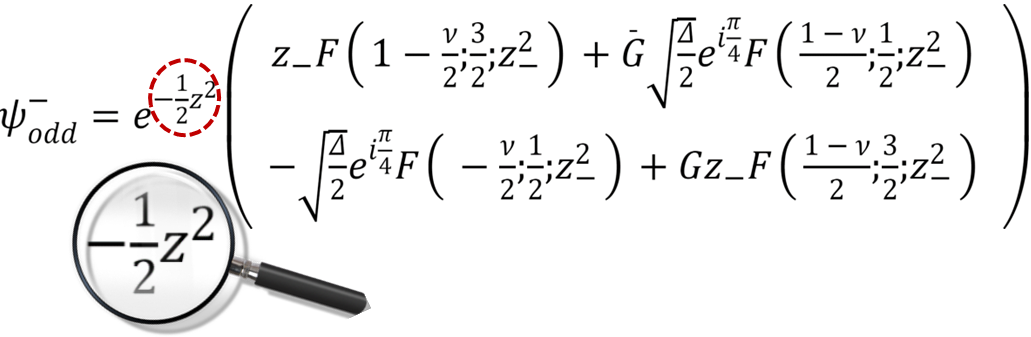}

    \caption{The model fails to capture intricate details in complex formulas, misrecognizing \(-\frac{1}{2}z^2\) as \(\frac{1}{2}z^2\).}
    \label{fig:motivation}
\end{figure}

Mathematical Expression Recognition (MER) \cite{DBLP:journals/ijdar/ChanY00} refers to the technology that automatically recognizes and parses mathematical formulas or expressions from Latex images \cite{wilkins1995getting}, handwritten notes, or text documents. This means that MER models need to have a deep understanding of mathematical expressions, including the order of operations, structural nesting, and operation precedence, especially for complex formulas. However, the current models are still not performing well in this area. This is not only limited by the lack of datasets with highly complex structures \cite{DBLP:journals/corr/abs-2404-15254, OleehyO_TexTeller}, but also by the numerous subtle details in complex formulas that models often overlook. Early research in MER mainly relied on traditional machine learning methods, which typically involved handcrafted feature extraction and pattern recognition techniques \cite{DBLP:journals/mta/KukrejaS22, DBLP:journals/pr/ChanY01}.  With the development of deep learning, MER algorithms based on Convolutional Neural Networks (CNNs) and the Transformer architecture \cite{vaswani2017attention} have been proposed. These methods have shown excellent performance in recognizing both simple mathematical expressions \cite{DBLP:conf/icdar/MahdaviZMVG19, Paruchuri_texify} and handwritten data \cite{DBLP:conf/eccv/LiYLLJBLB22, DBLP:journals/prl/LeIN19, DBLP:conf/aaai/BianQXLSW22}. However, they struggle with limited accuracy when processing complex and intricate mathematical expressions, especially in capturing the finer details of these formulas, shown in Figure~\ref{fig:motivation}.

To address these challenges, we introduce the HDR dataset, a large-scale resource for MER with more than one hundred million formulas, including the HDR-100M training set and the HDR-Test set, which covers a wide range of expression complexities. We also propose HDNet, an encoder-decoder-based MER framework with a hierarchical sub-formula module that improves accuracy in parsing complex formulas. The main contributions of this paper are:
\begin{itemize}
    \item \textbf{HDR dataset:} A large-scale, multi-label MER dataset, providing a robust foundation for model development and evaluation across diverse expressions.
    \item \textbf{HDNet:} A novel MER framework that combines an encoder-decoder structure with a hierarchical sub-formula module for subgraph detail enhancement, significantly improving formula parsing accuracy.
    \item \textbf{Fair evaluation:} An improved evaluation method designed to account for multiple valid interpretations of a formula, ensuring more equitable comparisons.
\end{itemize}

\section{Related Work}
\label{sec:format}
A popular approach to MER leverages CNNs. The ConvMath model \cite{DBLP:conf/icpr/YanZGYT20} combines an image encoder for feature extraction with a convolutional decoder for sequence generation, converting images of mathematical expressions into LaTeX format. The WAP \cite{DBLP:journals/pr/ZhangDZLHHWD17} model integrates CNNs and RNNs with attention to recognize handwritten expressions directly from 2D image layouts. Scale instability in mathematical expressions has been addressed by improving model performance across different scales through scale augmentation \cite{DBLP:conf/icfhr/LiJLZ20}.

Another MER algorithm is based on the Transformer architecture \cite{vaswani2017attention}. An Attention-Based Mutual Learning Network (ABM) has been introduced to generate LaTeX sequences from images using an attention mechanism \cite{DBLP:conf/aaai/BianQXLSW22}. UniMERNet \cite{DBLP:journals/corr/abs-2404-15254} is designed for the universality of mathematical expressions, enabling it to handle symbols and structures across languages, making it suitable for multilingual environments. Vary \cite{DBLP:journals/corr/abs-2312-06109} enhances document-level OCR efficiency by using a vocabulary network and a small decoder-only transformer for autoregressive visual vocabulary generation.

However, these models often struggle with complex structures, overlapping symbols, or poor image quality, leading to parsing errors.

\begin{figure*}[t]
  \centering
  \includegraphics[width=0.9\linewidth, height=6cm]
  {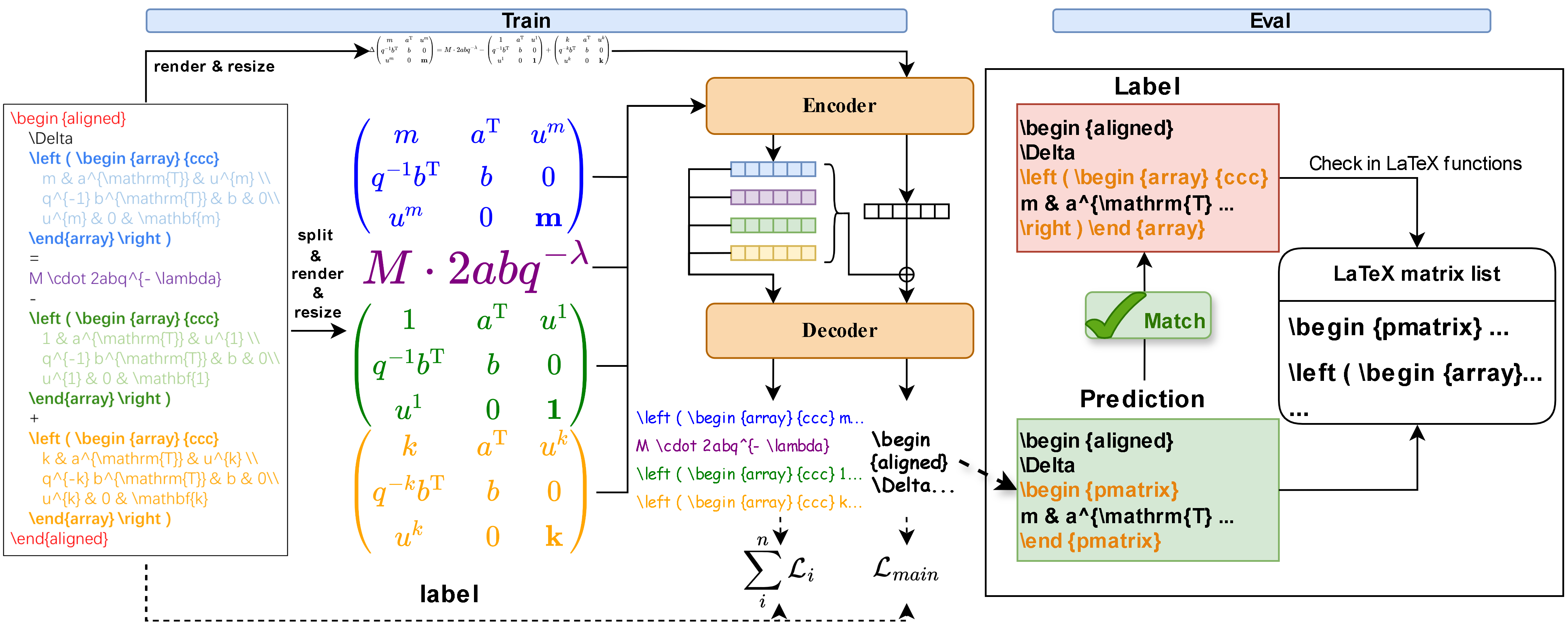}
  \caption{In the training process (left), the formulas are parsed hierarchically based on their labels. Each formula is split, rendered, and resized into sub-formulas. The main formula is also rendered and resized. Both the main formula and sub-formulas are fed into the encoder to extract features. The sub-formula features are then fused with the main formula's feature through weighted aggregation to provide additional visual details. The weighted features are passed to the decoder to predict the result for the main formula. Additionally, each sub-formula feature is separately passed to the decoder to predict sub-formula results. The model's optimization objective includes the loss of the main formula, $L_{\text{main}}$, and the sum of the losses of the sub-formulas, $\sum_i^n L_i$.The predicted results are evaluated (right). We provide a fair evaluation method where even if two formulas differ at the character level, they are considered correctly parsed if they are functionally equivalent.
}

  \label{method_main}
\end{figure*}

\section{HDR Dataset}
\label{sec:pagestyle}
We introduce the comprehensive HDR dataset, designed to tackle the challenges of recognizing complex mathematical formulas across disciplines, with a focus on explaining their hierarchical levels and complexity, as detailed in Section~\ref{Hierarchical}.

\subsection{Hierarchical levels}
\label{Hierarchical}
Hierarchical levels refer to the nested structures within mathematical formulas. At the simplest level (level 0), individual characters like \( a \) or \( 1 \) are standalone elements. Combining these characters with operations like superscripts or subscripts (e.g., \( a^2 \), \( b_1 \)) forms level 1 structures. More complex constructs, such as fractions (\( \frac{a}{b} \)) or summations (\( \sum_{i=1}^n a_i \)), belong to level 2. Higher levels arise from deeper nesting, such as matrices or piecewise functions (\(\begin{cases} 
a & \text{if } x > 0 \\ 
b & \text{otherwise} 
\end{cases}\)), often wrapped in \verb|\begin{...}\end{...}| environments.  

The challenge for models is accurately interpreting these nested relationships, ensuring that the opening and closing elements correspond correctly. For example, LaTeX matrices are wrapped within \verb|\begin{matrix}...\end{matrix}|, and the model must consider the entire structure to maintain coherence. Understanding these hierarchical levels is key to parsing complex formulas. Therefore, we designed the HDR dataset to include a range of formulas with varying hierarchical complexity, challenging models to accurately handle these structures while considering context.

\subsection{Data Collection and Statistic}
The HDR dataset comprises formulas extracted from arXiv PDFs (2007–2024), spanning fields like mathematics (math), statistics (stat), physics (phy), Quantitative Finance (q-fin), quantitative biology (q-bio), economics (econ), electrical engineering and systems science (eess), and computer science (cs)\footnote{\url{https://arxiv.org/category_taxonomy}}. This classification ensures broad domain coverage and diverse mathematical expression types, enhancing its relevance to various research areas.

We focus on capturing complex formulas and employ various synthesis techniques to enhance the dataset. This includes generating both inline and display formulas to ensure a comprehensive representation of mathematical expressions. Formula complexity is measured by hierarchical levels, lines, and characters. For example, matrices nested within fractions, or piecewise functions comprising multiple polynomials, they often are rendered across multiple lines. And the final HDR-100M contains 100,130,000 samples for train set and 43,932 for test. A key feature of HDR-Test is its one-to-many labeling, allowing multiple labels per formula to capture the diverse representations of expressions. Labels support hierarchical classification for flexible representation, shown in Table~\ref{datasets}.

\begin{table}[htbp]\scriptsize
    \caption{The statistics of multi-annotation HDR-Test set. The first row shows hierarchical levels of LaTeX formulas, including [1-2], [3-5] and [6-7]. The second row indicates the number of lines: A for [1-3], B for [4-8], C for [9-20], and D for [21-51].}
    \vspace{1em}
    \centering
    \begin{tabular}{p{0.6cm} p{0.2cm} p{0.2cm} p{0.2cm} p{0.2cm} | p{0.2cm} p{0.2cm} p{0.2cm} p{0.2cm} | p{0.2cm} p{0.2cm} p{0.2cm} p{0.2cm}}
    \toprule
    \textbf{Class} & \multicolumn{4}{c|}{\textbf{[1-2]}} & \multicolumn{4}{c|}{\textbf{[3-5]}} &
    \multicolumn{4}{c}{\textbf{[6-7]}}\\
    \cmidrule(r){2-5} \cmidrule(r){6-9} \cmidrule(r){10-13}
    & \textbf{A} & \textbf{B} & \textbf{C} & \textbf{D} & \textbf{A} & \textbf{B} & \textbf{C} & \textbf{D}  & \textbf{A} & \textbf{B} & \textbf{C} & \textbf{D}\\
    \midrule
    math & 2929 & 583 & 339 & 29 & 3181 & 2411 & 994 & 106 & 10 & 75 & 133 & 18\\ 
    stat & 1730 & 60 & 20 & 3 & 1629 & 453 & 158 & 18 & 3 & 15 & 13 & 19\\ 
    phy & 2119 & 282 & 99 & 7 & 4462 & 1693 & 615 & 60 & 31 & 56 & 49 & 21\\ 
    q-fin & 1423 & 19 & 19 & 0 & 978 & 174 & 48 & 8 & 3 & 4 & 7 & 0\\ 
    q-bio & 861 & 34 & 9 & 3 & 901 & 101 & 24 & 1 & 0 & 2 & 3 & 0\\ 
    econ & 2952 & 160 & 18 & 5 & 3281 & 490 & 111 & 15 & 13 & 9 & 8 & 0\\ 
    eess & 1670 & 36 & 4 & 0 & 911 & 167 & 56 & 15 & 2 & 17 & 3 & 0\\ 
    cs & 2648 & 201 & 100 & 4 & 1931 & 696 & 284 & 39 & 3 & 6 & 21 & 5\\ 
    \bottomrule
    \end{tabular}
    \label{datasets}
\end{table}

\section{FRAMEWORK}
\label{sec:method}
We propose the Hierarchical Detail Network (HDNet), to capture fine-grained hierarchical features in math formulas.

\subsection{HDNet Architecture}
When handling intricate formulas, fixed input image sizes can reduce resolution and lead to inaccuracies in recognizing fine details shown in Figure~\ref{method_main}. HDNet addresses  this with hierarchical decomposition and random cropping through its hierarchical sub-formula module, overcoming the fixed image size limitations in encoder-decoder models for complex formulas.
Hierarchical sub-formula module improves resolution and precision. It employs a sub-formula cropping strategy, decomposing formulas into sub-formulas rendered as high-resolution images. By using $n$ randomly selected sub-formula images, where their total character count is at least 70\% of the entire formula's, i.e., $\sum_i^n c_i \geq c$, HDNet captures essential details and focuses on intricate components, significantly improving recognition accuracy. During inference, sub-formulas cannot be extracted, and their labels are unavailable. To simulate this, random cropping of $\lambda\%$ of the samples is applied during training, ensuring the model's robustness during the prediction.

HDNet is built upon a Transformer-based encoder-decoder framework \cite{vaswani2017attention}. Each input formula image \( I \in \mathbb{R}^{3 \times H \times W} \) is cropped to generate a set of high-resolution sub-formula images, which are processed by a Vision Transformer \cite{DBLP:conf/iclr/DosovitskiyB0WZ21} encoder to produce feature vectors. Specifically, a feature vector for the main formula image \( Z_{\text{main}} \) and feature vectors for the sub-formula images \( Z_i \) are obtained. These vectors are then fused into a unified representation \( Z \), defined as:
\begin{equation}
    Z = \alpha \cdot Z_{\text{main}} + (1 - \alpha) \cdot \frac{1}{n} \sum_{i=1}^{n} Z_i
    \label{Zfeature}
\end{equation}
where \( \alpha \) balances the contributions of the main formula and sub-formula features, and \( n \) is the number of sub-formula images. The fused feature vector \( Z \) is first passed into the decoder. The decoder utilizes cross-attention mechanisms to interact with the output text sequence. It then generates the predicted formula from these interactions. This approach allows HDNet to integrate fine-grained details from both the main formula and sub-formulas for improved accuracy.

\subsection{Loss Function}
In addition to the formula decoding loss, sub-formula labels are used to compute the sub-formula loss, contributing to the overall optimization of the model. The overall optimization objective for HDNet is defined by the following loss function:

\begin{equation}
    \mathcal{L}_{total} = \alpha \cdot \mathcal{L}_{\text{main}} + (1 - \alpha) \cdot \frac{1}{n} \sum_{i=1}^{n} \mathcal{L}_{i}
    \label{eq:total_loss}
\end{equation}
where $\mathcal{L}_{\text{main}}$ is the primary loss of formula recognition. It is computed using the autoregressive language model loss:

\begin{equation}
    \mathcal{L} = - \sum_{t=1}^{T} \log p(y_t \mid y_{<t}, Z)
\end{equation}
where \( y_t \) denotes the token at position \( t \) in the output sequence, \( y_{<t} \) is the preceding tokens, and \( Z \) is the feature vector generated through Equation~\ref{Zfeature}. $\mathcal{L}_{i}$ is the loss for each sub-formula, capturing the intricate details of complex formulas. This sub-formula loss is included to balance the overall recognition of the entire formula with the detailed parsing of its components. The parameter $\alpha$ controls the trade-off between the primary formula recognition loss ($\mathcal{L}_{\text{main}}$) and the sub-formula losses ($\mathcal{L}_{i}$), ensuring a balanced optimization that improves both formula recognition accuracy and detailed parsing.

Through supervised parsing of formula details at a granular level, HDNet enhances the accuracy and robustness of MER models, effectively managing the complexities inherent in mathematical expressions.

\section{EXPERIMENTS}
\label{sec:majhead}

\subsection{Fair Evaluation Metrics}
Traditional evaluation methods, limited to the character level, often fail to fairly assess models when mathematical formula images can be interpreted in multiple valid ways. The predicted results and ground-truth labels may differ at the character level but render identically as images due to functionally equivalent LaTeX commands.

To overcome this, we propose a simple and efficient evaluation strategy that considers all valid parsing options for functionally equivalent expressions, as shown in Figure~\ref{method_main} (right). Our method replaces both the labels and model predictions with equivalent expressions before performing character-level evaluation. This approach accommodates a wider range of valid expressions and ensures a more robust and fair evaluation of model performance in recognizing and parsing complex formulas. By handling variability in LaTeX-based formula generation, our strategy enhances the accuracy and fairness of character-level evaluations.
The character-level evaluation metrics we use include average sample edit distance, BLEU score, and character recall. Character recall is defined as $CR = 1 - \frac{\text{Edit Distance}}{\text{Number of Characters}}$.

\subsection{Implementation Details}
\begin{figure}[h]
    \centering
    \includegraphics[width=0.45\textwidth]{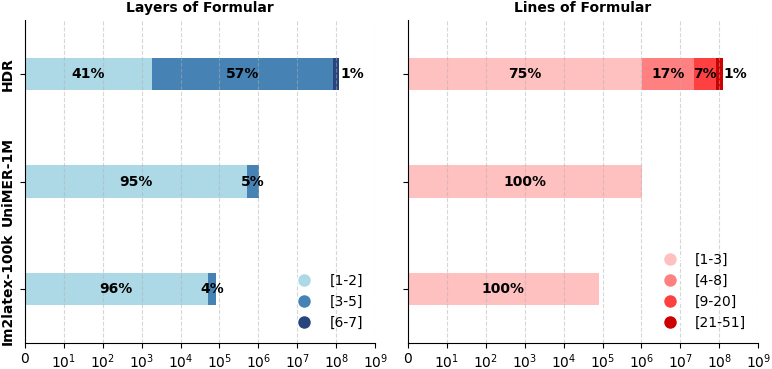}
    \caption{Comparison of datasets Im2latex-100k, UniMER-1M, and HDR, showing the number of hierarchical layers and the number of lines. Darker colors indicate higher complexity. The bar length represents total data volume.}
    \vspace{-\baselineskip}
    \label{fig2}
\end{figure}

\noindent\textbf{Datasets.}
We evaluated models on the HDR-Test dataset and two public datasets, Im2latex-100k and UniMER-1M, with varying complexity levels. The HDR dataset, as shown in Figure~\ref{fig2}, is the most complex and comprehensive.

\noindent\textbf{Experimental settings.} We used the OCR pretrained model TrOCR-base, with 12 Transformer layers in both encoder and decoder. The loss function combines main and sub-image losses (Eq.~\ref{eq:total_loss}) balanced by $\alpha=0.2$. The images were resized to 448×448 with $n=4$ sub-image segments per sample. The models were trained for 100 epochs with 32 batch size using AdamW (learning rate 1e-4) on 8 NVIDIA A100-80G GPUs.

\begin{table}[h]
\vspace{-\baselineskip}
\caption{Performance comparison of various models on the HDR-Test dataset, evaluated using Character Recall (CR), Average Edit Distance (AED), and BLEU Score (BS). NF stands for Non-Fair evaluation, and F stands for Fair evaluation. Higher CR and BS values (↑) indicate better performance, while lower AED values (↓) are preferable.}
\label{formula_comp}
\normalsize
\centering
\begin{tabular}{lcccccc}
\toprule
\textbf{Model} & \multicolumn{2}{c}{\textbf{ CR (↑)}} & \multicolumn{2}{c}{\textbf{ AED (↓)}} & \multicolumn{2}{c}{\textbf{BS (↑)}} \\
\cmidrule(r){2-3} \cmidrule(r){4-5} \cmidrule(r){6-7}
 & \textbf{NF} & \textbf{F} & \textbf{NF} & \textbf{F} & \textbf{NF} & \textbf{F} \\
\midrule
Pix2tex & 0.278 & 0.331 & 249 & 229 & 0.303 & 0.326 \\
Texify & 0.502 & 0.531 & 172 & 158 & 0.458 & 0.481 \\
UniMERNet & 0.574 & 0.618 & 147 & 125 & 0.559 & 0.615 \\
\midrule
\textbf{HDNet} & \textbf{0.952} & \textbf{0.968} & \textbf{16} & \textbf{13} & \textbf{0.925} & \textbf{0.931} \\
\bottomrule
\end{tabular}
\end{table}

\subsection{Overall Results}

We compared HDNet with Pix2tex, Texify, and UniMERNet, as shown in Table~\ref{formula_comp}. HDNet outperforms all baselines, achieving the highest character recall (0.952 non-fair, 0.968 fair), lowest average edit distances (16 non-fair, 13 fair), and highest BLEU scores (0.925 non-fair, 0.931 fair). These results demonstrate HDNet's effectiveness in accurately recognizing and reconstructing complex formulas.

\begin{table}[htbp]
\vspace{-\baselineskip}
\caption{
    Results of different models on the \textsc{Im2latex-100k} and \textsc{UniMER-1M} datasets. CR stands for Character Recall, and Fair-CR refers to Fair Character Recall.}
\label{pub_datasets}
\normalsize
\centering
\begin{tabular}{lcccccc}
\toprule
\textbf{Model} & \multicolumn{2}{c}{\textbf{Im2latex-100k}} & \multicolumn{2}{c}{\textbf{UniMER-1M}} \\
\cmidrule(r){2-3} \cmidrule(r){4-5}
 & \textbf{CR} & \textbf{Fair-CR} & \textbf{CR} & \textbf{Fair-CR} \\
\midrule
Pix2tex & 0.912 & 0.913 & 0.465 & 0.493 \\ 
Texify & 0.939 & 0.944 & 0.708 & 0.727 \\
UniMERNet & 0.942 & 0.948 & 0.908 & 0.911 \\
\midrule
\textbf{HDNet} & \textbf{0.979} & \textbf{0.982} & \textbf{0.951} & \textbf{0.963} \\
\bottomrule
\end{tabular}
\end{table}

\begin{figure}[h]
    \centering
    \includegraphics[width=0.38\textwidth] {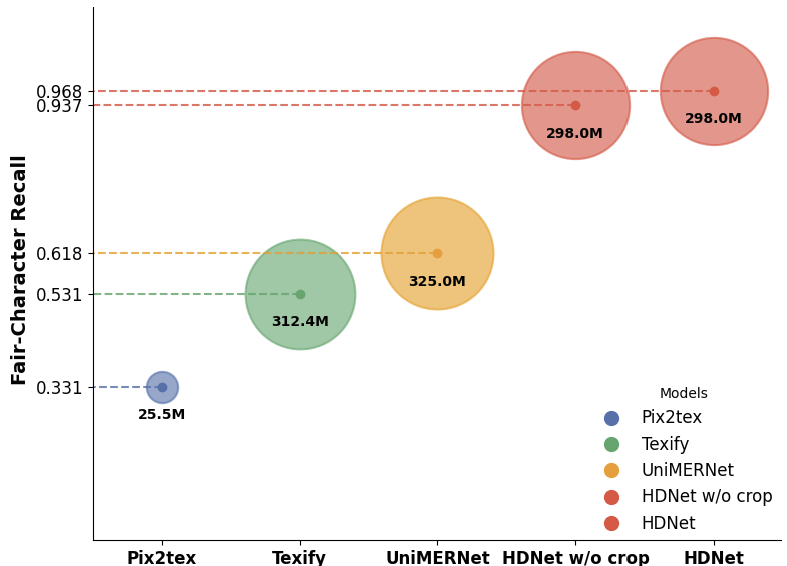}
    \caption{Comparison of different models based on parameter counts (represented by the area of circles) and Fair-Character Recall on the HDR dataset. Larger circles represent models with more parameters, while the vertical position reflects the Fair-Character Recall performance.}
    \vspace{-\baselineskip}
    \label{fig3}
\end{figure}

We evaluated HDNet on public datasets Im2latex-100k and UniMER-1M, as shown in Table~\ref{pub_datasets}. HDNet outperforms baselines, achieving 0.979 CR and 0.982 Fair-CR on Im2latex-100k, and 0.951 CR and 0.963 Fair-CR on UniMER-1M, surpassing UniMERNet in both metrics. These results highlight HDNet's superior accuracy and generalization. Notably, the improvements are achieved without increasing parameters, as the sub-formula method adds no extra complexity in Figure~\ref{fig3}.

\subsection{Ablation Studies}

To validate the hierarchical sub-formula module, we performed ablation experiments among 4 modules: no cropping, random cropping, sub-formula cropping, and a combination, as shown in Table~\ref{ablation_study}.

The best performance (Fair-CR score of 0.968) was achieved with the Sub-Formula + RandomCrop configuration. However, using Sub-FormulaCrop alone resulted in a lower Fair-CR score of 0.837 due to a mismatch between training and testing phases, as Sub-FormulaCrop relies on labels unavailable during testing. In the Sub-Formula + RandomCrop configuration, RandomCrop is applied during both training and testing. This approach eliminates mismatch, ensuring consistency and improved performance.

\begin{table}[htb]
\vspace{-1\baselineskip}
\caption{Ablation study comparing different croppting methods in formula recognition on HDR-Test set.}
\label{ablation_study}
\normalsize
\centering
\begin{tabular}{lcc}
\toprule
\textbf{Method} & \textbf{CR} & \textbf{Fair-CR} \\
\midrule
w/o Crop & 0.929 & 0.937 \\
Full Random Crop & 0.940 & 0.955 \\
Full Sub-Formula Crop & 0.816 & 0.837 \\
Sub-Formula + Random Crop & \textbf{0.952} & \textbf{0.968} \\
\bottomrule
\end{tabular}
\end{table}

\section{Conclusion}

This paper presents a subgraph framework to enhance model focus on mathematical expression details, introduces the HDR-100M dataset for MER research, and refines evaluation methods for fairer model comparisons.

\section*{Acknowledgement}
The authors thank their colleagues for their support.

\clearpage 
\bibliographystyle{IEEEbib}
\bibliography{main}
\end{document}